\documentclass[runningheads,a4paper]{llncs}

\usepackage{amssymb}
\usepackage{graphicx}
\usepackage{algorithm,algpseudocode}
\newtheorem{exmp}{Example}
\algnewcommand\algorithmicinput{\textbf{Input:}}
\algnewcommand\Input{\item[\algorithmicinput]}
\algnewcommand\algorithmicoutput{\textbf{Output:}}
\algnewcommand\Output{\item[\algorithmicoutput]}
\algnewcommand\algorithmicforeach{\textbf{for each}}
\algdef{S}[FOR]{ForEach}[1]{\algorithmicforeach\ #1\ \algorithmicdo}
\usepackage{graphicx}
\usepackage{enumitem}
\usepackage{booktabs}

\usepackage{url}
\urldef{\mailsa}\path|{fxs130430,gupta@utdallas.edu|

\begin{document}

\mainmatter  

\title{Induction of Non-Monotonic Logic Programs to Explain Boosted Tree Models Using LIME}

\titlerunning{Lecture Notes in Computer Science: Authors' Instructions}

%
%
\author{Farhad Shakerin \and Gopal Gupta}
\authorrunning{F. Shakerin and G. Gupta}

\institute{Computer Science Department \\ The University of Texas at Dallas, Richardson, USA
\email{\{fxs130430,gupta\}@utdallas.edu}}
%
%

\toctitle{ILP to Explain XGBoost models Using LIME}
\tocauthor{F. Shakerin and G. Gupta}
\maketitle

\begin{abstract}
We present a heuristic based algorithm to induce \textit{nonmonotonic} logic programs that will explain the behavior of XGBoost trained classifiers. We use the technique based on the LIME approach to locally select the most important features contributing to the classification decision. Then, in order to explain the model's global behavior, we propose the LIME-FOLD algorithm ---a heuristic-based inductive logic programming (ILP) algorithm capable of learning non-monotonic logic programs---that we apply to a transformed dataset produced by LIME. Our proposed approach is agnostic to the choice of the ILP algorithm. Our experiments with UCI standard benchmarks suggest a significant improvement in terms of classification evaluation metrics. Meanwhile, the number of induced rules dramatically decreases compared to ALEPH, a state-of-the-art ILP system.    
\end{abstract}

\section{Introduction}
Dramatic success of machine learning has led to a torrent of Artificial Intelligence (AI) applications. However, the effectiveness of these systems is limited by the machines' current inability to explain their decisions and actions to human users. That's mainly because the statistical machine learning methods produce models that are complex algebraic solutions to optimization problems such as risk minimization
or data likelihood maximization. Lack of intuitive descriptions makes it hard for users to understand and verify the underlying rules that govern the model. Also,
these methods cannot produce a justification for a prediction
they compute for a new data sample.

The Explainable AI program \cite{xai} aims to create a suite of machine learning techniques that: a) Produce more explainable models, while maintaining a high level of prediction accuracy. b) Enable human users to understand, appropriately trust, and effectively manage the emerging generation of artificially intelligent partners. 

Inductive Logic Programming (ILP) \cite{ilp} is one Machine Learning technique where the learned model is in the form of logic programming rules (Horn Clauses) that are comprehensible to humans. It allows the background knowledge to be incrementally extended without requiring the entire model to be re-learned. Meanwhile, the comprehensibility of symbolic rules makes it easier for users to understand and verify induced models and even edit them. 

The ILP learning problem can be regarded as a search problem for a set of clauses that deduce the training examples. The search is performed either top down or bottom-up. A bottom-up approach builds most-specific clauses from the training examples and searches the hypothesis space by using generalization. This approach is not applicable to large-scale datasets, nor it can incorporate \textit{Negation-As-Failure} into the hypotheses. A survey of bottom-up ILP systems and their shortcomings can be found at \cite{sakama05}. In contrast, top-down approach starts with the most general clauses and then specializes them. A top-down algorithm guided by heuristics is better suited for large-scale and/or noisy datasets \cite{quickfoil}.

The FOIL algorithm by Quinlan \cite{foil} is a popular top-down algorithm. FOIL uses  heuristics from information theory called \textit{weighted information gain}. The use of a greedy heuristic allows FOIL to run much faster than bottom-up approaches and scale up much better. For instance, the QuickFOIL system \cite{quickfoil} can deal with millions of training examples in a reasonable time. However, scalability comes at the expense of losing accuracy if the algorithm is stuck in local optima and/or when the number of examples is insufficient. The former is an inherent problem in hill climbing search and the latter is due to the shrinking of examples during clause specialization. Also, elimination of already covered examples from the training set (to guarantee the termination of FOIL) causes a similar impact on the quality of heuristic search for the best clause. Therefore, the predicates picked-up by FOIL are not always globally optimal with respect to the concept being learned. 
Based on our research, we believe that a successful ILP algorithm must satisfy the following criteria:

\begin{itemize}
\item It must employ heuristic-based search for clauses for the sake of scalability.
\item It should be able to figure out  relevant features, regardless of the number of current training examples.
\item It should be able to learn from incomplete data, as well as be able to distinguish between noise and exceptions.
\end{itemize}

Unlike top-down ILP algorithms, statistical machine learning methods are bound to find the relevant features because they optimize an objective function with respect to global constraints. This results in models that are inherently complex and cannot explain what features  account for a classification decision on any given data sample. 

Recently, some solutions have been proposed by researchers to explain black-box classifiers' predictions locally. LIME \cite{lime} is a novel model-agnostic system that explains the classification decisions made by any classifier on any given data sample. The idea comes from the fact that explaining classifier's behavior in a local region around any data turns out to be easier than explaining its global behavior. Each local explanation is a set of feature-value pairs that would determine what features and how strongly each feature, relative to other features, contributes to the classification decision.

In order to capture model's global behavior, we propose an algorithm called LIME-FOLD, to learn concise logic programs from a transformed data set that is generated by storing the explanations provided by LIME. The LIME system takes 
as input
a black-box model (such as a Neural Network, Random Forest, etc.) and a data sample. For any given data sample, it outputs a list of (weighted) features that contribute most to the classification decision. By repeating the same process for all training samples, we can generate a transformed version of the original data set that only contains the relevant features for each data sample. 

The LIME-FOLD algorithm learns a non-monotonic logic program from the transformed data set. This logic program explains the global behavior of the model. Our experiments on 10 UCI standard benchmark suggests that the hypotheses generated by LIME-FOLD algorithm are very concise and outperform the baseline ALEPH system \cite{aleph}. It also outperforms ALEPH once ALEPH is given the transformed dataset (i.e., ALEPH is extended with the LIME technique). Performance is measured in terms of classification evaluation scores, number of generated clauses and running time. 

Although LIME is model-agnostic, in this research we incorporate the XGBoost algorithm to train our statistical models. XGBoost \cite{xgboost} is a scalable tree boosting machine learning algorithm that is widely used by data scientists to achieve state-of-the-art results on many challenges. In essence, the hypotheses (a nonmonotonic logic program) that our LIME-FOLD algorithm induces, explain the behavior of XGBoost models.

This paper makes the following novel contribution: We present a new ILP algorithm capable of learning non-monotonic logic programs from local explanations of boosted tree models provided by LIME. We call this new algorithm LIME-FOLD. The LIME-FOLD algorithm is a scalable heuristic-based algorithm that explains the behavior of boosted tree models globally and outperforms ALEPH in terms of classification evalutation metrics as well as in providing more  concise explanations measured in terms of number of clauses induced.

\section{Background}
\label{sec:background}
\subsection{Problem Definition}
Inductive Logic Programming (ILP) \cite{ilp} is a subfield of machine learning that learns models in the form of logic programming rules (Horn Clauses) that are comprehensible to humans. This problem is formally defined as:\\
\textbf{Given}
\begin{enumerate}
    \item a background theory $B$, in the form of an extended logic program, i.e., clauses of the form $h \leftarrow l_1, ... , l_m,\ not \ l_{m+1},...,\ not \ l_n$, where $l_1,...,l_n$ are positive literals and \textit{not} denotes \textit{negation-as-failure} (NAF) and $B$ has no even cycle
    \item two disjoint sets of ground target predicates $E^+, E^-$ known as positive and negative examples respectively
    \item a hypothesis language of function free predicates $L$, and a  refinement operator $\rho$ under $\theta-subsumption$ \cite{plotkin70} that would disallow even cycles.
\end{enumerate}
\textbf{Find} a set of clauses $H$ such that:
\begin{itemize}
    \item $ \forall e \in \ E^+ ,\  B \cup H \models e$
    \item $ \forall e \in \ E^- ,\  B \cup H \not \models e$
    \item $B \land H$ is consistent.
\end{itemize}
\subsection{The FOIL Algorithm}
The LIME-FOLD algorithm is an extension of the FOIL algorithm \cite{foil}. Therefore, we first briefly discuss the FOIL algorithm. FOIL is a top-down ILP algorithm which follows a \textit{sequential covering} approach to induce a hypothesis. The FOIL algorithm is summarized in Algorithm \ref{algo:foil}. This algorithm repeatedly searches for clauses that score best with respect to a subset of 
positive and negative examples, a current hypothesis and a heuristic called \textit{information gain} (IG). 

\begin{algorithm}
\caption{Summarizing the FOIL algorithm}
\label{algo:foil}
\begin{algorithmic}[1]
\Input $target,B,E^+,E^-$ 
\Output 
Initialize $H \gets \emptyset $
\While{($|E^+| > 0$)}
	\State $c \gets (target$ :- $ \ true.)$
	\While{($|E^-| > 0 \land c.length < max\_length $)}
		\For{all $ \ c' \in \rho (c)$}
        	\State $compute \ score(E^+,E^-,H \cup \{c'\},B)$
    	\EndFor
    	\State let $\hat{c}$ be the $c' \in \rho(c)$ with the best score   
         \State $E^- \gets covers(\hat{c},E^-)$
    \EndWhile	
    \State add $\hat{c}$ to $H$
    \State $E^+ \gets E^+ \setminus covers(\hat{c},E^+)$
\EndWhile 
\State \textbf{return} $H$
\end{algorithmic}
\end{algorithm}

The inner loop searches for a clause with the highest information gain using a general-to-specific hill-climbing search. To specialize a given clause $c$, a refinement operator $\rho$ under $\theta$-subsumption ~\cite{plotkin70} is employed. The most general clause is $p(X_1,...,X_n) \gets true.$ where the predicate $p/n$ is the predicate being learned and each $X_i$ is a variable. The refinement operator specializes the current clause $h \gets b_1,...b_n .$ This is realized by adding a new literal $l$ to the clause yielding $h \gets b_1,...b_n,l$. The heuristic based search uses information gain. In FOIL, information gain for a given clause is calculated as follows: 
\begin{equation}
IG(L,R) = t\left(log_2 \frac{p_1}{p_1 + n_1} - log_2 \frac{p_0}{p_0+ n_0} 
\right)
\end{equation}
where $L$ is the candidate literal to add to rule $R$, $p_0$ ($n_0$) is the number of 
positive (negative) examples covered by $R$ respectively, $p_1$ ($n_1$) is the number of positive (negative) examples covered by $R+L$ respectively, and $t$ is the number of positive examples that are covered by $R$ and $R+L$ together. 

\subsection{The LIME Approach}
LIME \cite{lime} is a novel technique that finds easy to understand explanations for the predictions of any complex black-box classifier in a faithful manner. LIME constructs a linear model by sampling $N$ instances around any given data sample $x$. Every instance $x'$ represents a perturbed version of $x$ where perturbations are realized by sampling uniformly at random for each feature of $x$. LIME stores the classifier decision $f(x')$ and the kernel $\pi(x,x')$. The  $\pi$ function measures how similar the original and perturbed sample are and it is then used as the associated weight of $x'$ in fitting a \textit{locally weighted linear regression} (LWR) curve around $x$. The $K$ greatest learned weights of this linear model are interpreted as top $K$ contributing features into the decision made by the black-box classifier. Algorithm \ref{algo:lime} illustrates how a locally linear model is created around $x$ to explain a classifier's decision.

\begin{figure}
    \includegraphics[width=\textwidth]{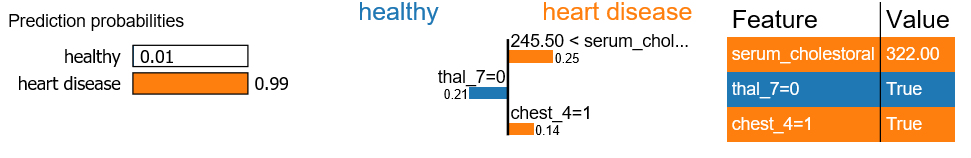}
\caption{Top 3 Relevant Features in Patient Diagnosis According to LIME}
\label{fig:heartlime}
\end{figure}

\begin{algorithm}
\caption{Linear Model Generation by LIME}
\label{algo:lime}
\begin{algorithmic}[1]
\Input $f:$ Classifier
\Input $N:$ Number of samples, $K:$ length of explanation,
\Input $x:$ sample to explain, $\pi :$ similarity kernel
\Output $w:$ fitted curve's weights  
\State $\mathcal{Z} \gets \{\}$
\For{$ i \in \{1,2,3,...,N\}$}
    \State // $x'_i$ is generated by perturbing features of $x$
	\State $x'_i \gets sample\_around(x)$
	\State $\mathcal{Z} \gets \mathcal{Z} \cup \langle x'_i,f(x'_i),\pi(x'_i,x) \rangle$ 
\EndFor
\State  // Fit a line to (weighted) points in $\mathcal{Z}$
\State $w \gets LWR(\mathcal{Z},K)$
\State \textbf{return} $w$
\end{algorithmic}
\end{algorithm}
\noindent
The interpretation language should be understandable by humans. Therefore, LIME requires the user to provide some interpretation language as well. In case of tabular data, it boils down to specifying the valid range of each table column. In particular, if the data column is a numeric variable (as opposed to categorical), the user must specify the intervals or a discretization strategy to allow LIME to create intervals that are used later on to explain the classification decision.
\begin{exmp}
\label{ex:heart}
The UCI heart dataset contains features such as patient's  blood pressure, chest pain, thallium test results, number of major vessels blocked, etc. The classification task is to predict whether the subject suffers from heart disease or not. Figure \ref{fig:heartlime} shows how LIME would explain a model's prediction over a data sample.   
\end{exmp}
In this example, LIME is called to explain why the model predicts heart disease. In response, LIME returns the top features along with their importance weight. According to LIME, the model predicts ``heart disease" because of high serum cholesterol level, and having a chest pain of type 4 (i.e., asymptomatic). In this dataset, chest pain level is a categorical variable with 4 different values.

The categorical variables should be \textit{binarized} before a statistical model can be applied. Binarization is the process of transforming each categorical variable with domain of cardinality $n$, into $n$ new binary features. The feature ``thallium test'' is a categorical feature too. However, in this case LIME reports that the feature ``thal\_7" which is a new feature that resulted from binarization and has the value ``false'', would have made the model predict ``healthy". The value 7 for thallium test in this dataset indicates reversible defect which is a strong indication of heart disease. It should be noted that the feature ``serum cholesterol" is discretized with respect to the training examples' label. Discretization aims to reduce the number of values a continuous variable takes by grouping them into intervals. Discretization method should maximize the interdependence between the variable values and the class labels. One of the most practiced methods for discretizing continuous data is the MDL method \cite{fayyadirani} which uses mutual information to recursively define the best bins. In this research, we discretize all numeric features using the MDL method. 

\section{Learning Default Theories}
\label{sec:Fold}
ILP algorithms such as FOIL induce logic programs that contain negated goals of the form \textit{not p}, where the \textit{not} is considered as \textit{negation as failure}. Logic programs containing negation-as-failure (with nonmonotonic semantics) are more expressive and concise when it comes to describing concepts (e.g., default reasoning, used in common sense reasoning used by humans) \cite{baral}. The handling of negation-as-failure in the FOIL algorithm is problematic. We illustrate these problems in \cite{fold} with many compelling examples. In addition, \cite{fold} builds upon the FOIL algorithm to develop a new algorithm called FOLD (First Order Learner of Defaults) for inducing default theories. FOLD induces nonmonotonic logic programs that are more precise and more concise. In this paper we build upon our work in \cite{fold}: we adapt and integrate the FOLD algorithm with LIME to develop a new algorithm called LIME-FOLD that not only is more accurate, it also produces significantly more concise nonmonotonic logic programs as explanations.

The FOLD algorithm which is an extension of FOIL, learns a concept in terms of a default and possibly multiple exceptions (and exceptions to exceptions, exceptions to exceptions of exceptions, and so on). FOLD tries first to learn the default by specializing a general rule of the form \texttt{\{target($V_1,...,V_n$) :- true.\}} with positive literals. 
As in FOIL, each specialization must rule out some already covered negative 
examples without decreasing the number of positive examples covered 
significantly. Unlike FOIL, no negative literal is used at this stage. Once 
the heuristic score (i.e., \textit{information gain}) (IG) becomes zero, or the maximum clause length is reached (whichever happens first), this process stops. At this point, if any negative 
example is still covered, they must be either noisy data or 
exceptions to the current hypothesis. Exceptions could be learned by swapping the current positive and negative examples, then calling the same algorithm recursively. As a result of this recursive process, FOLD can learn exceptions to exceptions, and so on. In presence of noise, FOLD identifies and \textit{enumerates} noisy samples, that is, outputs them as ground facts in hypothesis, to make sure that the algorithm converges. \textit{Maximum Description Length Principle} \cite{foil} is incorporated to heuristically control the hypothesis length and identify noise. Algorithm \ref{algo:fold} presents FOLD's pseudo-code.
\begin{algorithm}[!h]
\caption{FOLD Algorithm}
\label{algo:fold}
\begin{algorithmic}[1]
\Input $target, B,E^+,E^-$ 
\Output  $D = \{c_1,...,c_n\}$ \Comment{defaults' clauses}
		 \Statex $AB = \{ab_1,...,ab_m\}$ \Comment{exceptions/abnormal 
clauses}
\Function{FOLD}{$E^+,E^-$} 
\While{($|E^+| > 0$)}
\State $c \gets (target$ :- $ \ true.)$
\State $\hat{c} \gets$ \Call{specialize}{{c},{$E^+$},{{$E^-$}}}
\State $E^+ \gets E^+ \setminus 
covers(\hat{c},E^+,B)$
\State $D \gets D \cup \{ \hat{c} \}$
\EndWhile 
\EndFunction
\Function{SPECIALIZE}{${c},{E^+},{E^-}$}
\While{$|E^-|>0 \land c.length < max\_rule\_length$}
\State  $(c_{def},\hat{IG}) \gets$ \Call{add\_best\_literal}{{c},{$E^+$},{{$E^-$}}}
\If{$\hat{IG} > 0$}
	\State $\hat{c} \gets c_{def} $
\Else
	\State $\hat{c} \gets \Call{exception}{{c},{E^-},{{E^+}}}$
			 \If {$\hat{c} == null$}
						\State $\hat{c} \gets enumerate(c,E^+)$
			\EndIf	
\EndIf
\State $E^+ \gets E^+ \setminus 
covers(\hat{c},E^+,B)$
\State $E^- \gets covers(\hat{c},E^-,B)$
\EndWhile
\EndFunction

\Function{EXCEPTION}{${c_{def}},{E^+},{E^-}$}
\State  $\hat{IG} \gets 
\Call{add\_best\_literal}{{c},{E^+},{{E^-}}}$
\If{$\hat{IG} > 0$}
	\State $ c\_set \gets \Call{FOLD}{E^+,E^-} $
	\State $ c\_ab \gets generate\_next\_ab\_predicate()$
	\ForEach {$c \in c\_set $}
		\State $AB \gets AB \cup \{ c\_ab $:-$ \ bodyof(c) \}$
	\EndFor
	\State $\hat{c} \gets (headof(c_{def}) $:-$ \ bodyof(c), 
\textbf{not}(c\_ab))$
\Else
	\State $\hat{c} \gets null$
\EndIf

\EndFunction
\end{algorithmic}
\end{algorithm}
\begin{exmp}
\label{ex:pinguin}
$ B, E^+, E^-$ are background knowledge, positive and negative examples 
respectively. The target, i.e., the predicate being learned is \texttt{fly(X)}.
\end{exmp}
\begin{verbatim}
B:  bird(X) :- penguin(X).
    bird(tweety).   bird(et).
    cat(kitty).     penguin(polly).
E+: fly(tweety).    fly(et).
E-: fly(kitty).     fly(polly).
\end{verbatim}
By calling FOLD, at line 2 while loop, the clause 
\texttt{\{fly(X) :- true.\}} is specialized. Inside the $SPECIALIZE$ function, 
at line 10, the 
literal \texttt{bird(X)} is selected to add to the current clause, to get the 
clause 
$\hat{c}$ = \texttt{fly(X) :- bird(X)}, which happens to have the greatest IG 
among \texttt{\{bird,penguin,cat\}}. Then, at lines 20-21 the following updates 
are 
performed: $E^+=\{\}$,\ $E^-=\{polly\}$. A negative example 
$polly$, a penguin is still covered. In the next iteration, $SPECIALIZE$ fails 
to introduce a positive literal to rule it out since the best IG in this case 
is zero. Therefore, the EXCEPTION function is called by swapping the 
$E^+$, $E^-$. Now, FOLD is recursively called to learn a 
rule for $E^+ = \{polly\}$, $E^-=\{\}$. The recursive call 
(line 27), returns \texttt{\{fly(X) :- penguin(X)\}} as the exception. In line 
28, 
a new predicate \texttt{ab0} is introduced and at lines 29-31 the clause 
\texttt{\{ab0(X) :- penguin(X)\}} is created and added to the set of invented 
abnormalities, namely, AB. In line 32, the negated exception (i.e \texttt{not 
ab0(X)}) and the default rule's body (i.e \texttt{bird(X)}) are compiled 
together to form the following theory:
\begin{center}
    \begin{tabular}{l}
        \texttt{fly(X) :- bird(X), not ab0(X).}\\     
        \texttt{ab0(X) :- penguin(X).}     
    \end{tabular}
\end{center}

Once the FOLD algorithm terminates and a hypothesis is created, it would iterate through each clause's body and eliminates the redundant or counterproductive predicates. These are the predicates whose elimination does not make the clause cover significant number of negative examples. Next, FOLD sorts the hypothesis clauses in ascending order based on the number of positive examples each clause covers. Then, starting from the smallest, FOLD eliminates each clause and measures the coverage of positive examples. If elimination of a clause does not affect the overall coverage, the clause is removed from the hypothesis permanently.

\subsection{The LIME-FOLD Algorithm}
\label{sec:fold+lime}
In this section we introduce the LIME-FOLD algorithm by integrating FOLD and LIME. This yields a powerful ILP algorithm capable of learning very concise logic programs from a transformed dataset. The new algorithm outperforms FOLD and ALEPH \cite{aleph} which is a state-of-the-art ILP system. 

There are two major issues with the \textit{sequential covering} algorithms such as FOIL (and FOLD): 1) As number of examples decreases during specialization loop, probability of introducing an irrelevant predicate that accidentally splits a particular set of examples increases.  2) elimination of positive examples that are covered in previous iterations, impacts the precision of heuristic scoring. By filtering out the irrelevant features of each training example, the greedy clause search procedure is forced to pick up predicates only from a relevant subset of features to cover training examples. Relevant features for each training example is found by LIME once it is given an accurate classifier.

For instance in Figure \ref{fig:heartlime}, for a particular training example with 13 features, LIME returns only 3 as relevant to the underlying concept of heart disease on that particular data sample. This helps the FOLD algorithm to always pick up the relevant features regardless of the number of examples left. 

The success of this approach highly depends on the choice of statistical algorithm as well as tuning its parameters to make sure that the model makes the fewest errors in its predictions. In this research we conducted all experiments using the ``Extreme Gradient Boosting" (XGBoost) algorithm \cite{xgboost}. XGBoost is an implementation of the Gradient Boosted Decision Tree algorithm. Although LIME is model agnostic, in the experiments presented in this paper, XGBoost happened to always lead to better results.  

Algorithm \ref{algo:datatransformation} shows how a standard tabular dataset is transformed into an ILP problem instance for the FOLD algorithm. This algorithm takes a dataset $DS$, a target predicate $t$, and a classifier model $M$ that takes a feature vector and returns a binary classification value from the set \{'+','-'\}. For all data rows $r$ in $DS$, there is an identifier that is denoted by $r.id$. The numeric features once discretized are sorted based on the produced intervals and the interval index in the sorted list is used as the second argument of such features in generating the background knowledge. For instance let a numeric feature such as blood pressure be discretized first and stored as a sorted list of intervals as follows: $\{(-\infty,97),[97,120),[120,153),[153,170),[170,+\infty)\}$. The corresponding predicate for the datarow $r$ with $r.id = 135$ and blood pressure value 130 is \texttt{blood\_pressure(135,2)} because $135 \in [120,153)$ whose index in the above list is 2.

\begin{algorithm}
\caption{Dataset Transformation with LIME}
\label{algo:datatransformation}
\begin{algorithmic}[1]
\Input $t:$ target predicate,$DS:$ Dataset
\Input $M:$ trained classifier
\Output $BK:$ background knowledge
\Output $E^+$,$E^-: $ positive and negative examples
\State propositionalize categorical features
\State discretize numeric features
\ForEach{$DataRow \ r \in DS$}
	\If{$M(r) = $'+'}
    	\State $E^+ = E^+ \cup \{t(r.id)\}$
    \Else 
    	\State $E^- = E^- \cup \{t(r.id)\}$
    \EndIf
	\State explanation = LIME(M,r)
    \ForEach{$pair(e,w) \in explanation$}
    	\If{e is the $n^{th}$ discretized interval feature f}
        	\State $BK = BK \cup \{f(r.id,n)\}$ 
        \EndIf
    	\If{$e$ is an equality expr. of the form $f_v = 0$}
    		\State  // `-' denotes classical negation
        	\State $BK = BK \cup \{$-$f(r.id,v)\}$ 
        \EndIf
        \If{ $e$ is an equality expr. of the form $f_v = 1$} 
        	\State $BK = BK \cup \{f(r.id,v)\}$
        \EndIf
    \EndFor
\EndFor
\end{algorithmic}
\end{algorithm}
In Algorithm \ref{algo:datatransformation}, explanation pairs with negative weights are retrieved too. These are the features that would turn the classification decision into the opposite of concept we are learning. For instance in Example \ref{ex:heart}, a healthy subject may happen to have a high level ``serum cholesterol''. Therefore, if LIME-FOLD algorithm picks up this feature to cover some positive examples, the healthy subjects---which are negative examples---are also covered. The LIME-FOLD algorithm is able to rule these negative examples out by introducing an abnormality predicate that would make use of these negative weighted features. These are the features that led the XGBoost model to predict those subjects as healthy. 

\section{Experiments}
\label{sec:Experiments}
In this section, we present our experiments on UCI standard benchmarks \cite{uci}. The ALEPH system \cite{aleph} is used as the baseline. ALEPH is a state-of-the-art ILP system that has been widely used in prior work. To find a rule, ALEPH starts by building the most specific clause, which is called the ``bottom clause",
that entails a seed example. Then, it uses a branch-and-bound
algorithm to perform a general-to-specific heuristic search for a subset of literals from the bottom clause to form a more general rule. We set ALEPH to use the heuristic
enumeration strategy, and the maximum number of branch nodes to be explored in a branch-and-bound search to 500K. We use the standard metrics including precision, recall, accuracy and $F_1$ score to measure the quality of the results. We separately report the running time comparison as well.
\begin{figure}
    \includegraphics[width=\textwidth]{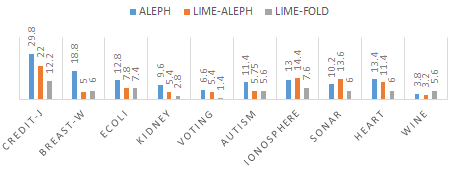}
\caption{Average Number of Rules Induced by Each Different Experiment}
\label{fig:numberofrules}
\end{figure}
We conduct three different sets of experiments as follows: First, we run ALEPH on 10 different datasets using 5-fold cross-validation setting. Second, each dataset is transformed as explained in Algorithm \ref{algo:datatransformation}. Then the LIME-FOLD algorithm is run on a 5-fold cross-validated setting, and the classification metrics are reported. Third, ALEPH is run on the same datasets produced in the second experiment. We call this approach LIME-ALEPH. 

Figure \ref{fig:numberofrules} compares the average number of clauses generated by standard ALEPH, LIME-ALEPH and LIME-FOLD on 10 UCI datasets. With the exception of ``breast-w" and ``wine'', in all other datasets, LIME-FOLD discovers fewer number of clauses. However, in ``breast-w'' and ``wine" the $F_1$ score of LIME-FOLD is higher than two other approaches. Also, it is worth noting that LIME-ALEPH in most cases generates fewer clauses than ALEPH. However, incorporating \textit{Negation-As-Failure} in LIME-FOLD algorithm as well as learning the clauses in terms of defaults and exceptions allows the algorithm to cover all positive examples with fewer number of clauses. 

Another observation that explains the advantage of LIME-ALEPH over ALEPH, is that LIME is capable of explaining propositionalized categorical variables in both affirmative and negative ways. For instance, in the ``UCI heart'' dataset, the \textit{thallium-201 stress scintigraphy test} is a categorical feature with three possible values in the set \{3,6,7\}, indicating normal, fixed defect and reversible defect in that order. The covering approach incorporated in ALEPH, would come up with two clauses corresponding to 6, 7, whereas, in both LIME-ALEPH and LIME-FOLD a negated feature \textit{f $ \neq $ 3} is introduced and stored in the transformed dataset.     

The following logic program is induced by LIME-FOLD algorithm (using the entire data set):
{\scriptsize
\begin{verbatim}
(1) heart_disease(A):- chest_pain(A,4), -thal(A,3).
(2) heart_disease(A):- slope(A,2), major_vessels(A,1).
(3) heart_disease(A):- chest_pain(A,4), sex(A,1),
                       not ab0(A).
(4) heart_disease(A):- blood_pressure(A,5), sex(A,1).
(5) heart_disease(A):- slope(A,2), blood_pressure(A,5).
(6) heart_disease(A):- slope(A,2), major_vessels(A,3),
                       serum_cholestoral(A,3).
    ab0(A):-major_vessels(A,3).
\end{verbatim}  
}
The induced program can be understood as follows: In clause (1), \texttt{chest\_pain(A,4)} indicates an asymptomatic type of chest pain. While \texttt{thal(A,3)} would indicate a \textit{thallium test} with normal results, the classically negated predicate \texttt{-thal(A,3)} indicates a proof that the \textit{thallium test} is abnormal. In clause (2) \texttt{slope(A,2)} indicates \textit{the slope of the peak exercise relative to rest} is flat, which is an indication of heart disease. \texttt{major\_vessels(A,N)} indicates a patient with $N$ (range: 0-3) colored major vessels during a Fluoroscopy test. The higher the number, the less narrowed major vessels. Clause (3) introduces an abnormality predicate which stipulates that an asymptomatic chest pain is an indication of heart disease unless there are no narrowed major vessels. High cholestrol and High blood pressure are specified in discretized intervals represented by their index. For instance \texttt{serum\_cholestoral(A,3)} denotes the cholesterol range between 245 mg/dl and 400 mg/dl in this dataset. Similarly, \texttt{blood\_pressure(A,5)} indicates systolic range between 15.7 and 18.6. 

\begin{figure}
    \centering
    \includegraphics[width=\textwidth]{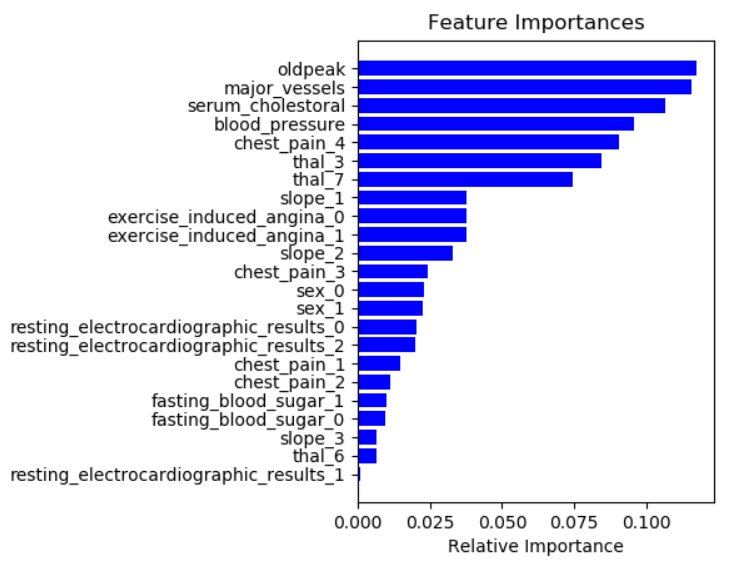}
\caption{XGboost Feature Importance Plot for UCI Heart}
\label{fig:globalfeatureimportance}
\end{figure}

Figure \ref{fig:globalfeatureimportance} shows the ``feature importance" plot calculated by xgboost algorithm. Generally, importance provides a score that indicates how useful or valuable each feature was in the construction of the boosted decision trees within the model. The more an attribute is used to make key decisions with decision trees, the higher its relative importance. Importance is calculated for a single decision tree by the amount that each attribute split point improves the performance measure, weighted by the number of observations the node is responsible for. The LIME-FOLD approach, prefers the more ``important" features over less ``important" ones, because the weighted information gain heuristic scores clauses with more frequently used features higher. ALEPH induces 18 clauses on the same data. Some of the features that the plot reports as rarely used by xgboost to split a node are introduced by ALEPH which makes the theory less relevant compared to what LIME-FOLD induces. 

\begin{table*}[]
\centering
\begin{tabular}{l|c|c|c|c|c|c|c|c|c|c|c|c|}
\cline{2-13}
                                 & \multicolumn{12}{c|}{Algorithm}                                                                                                                                                                                                                                                                                                                        \\ \hline
\multicolumn{1}{|l|}{Data Set}   & \multicolumn{4}{c|}{Aleph}                                                                                       & \multicolumn{4}{c|}{Aleph+Lime}                                                                                  & \multicolumn{4}{c|}{\textbf{Fold+Lime}}                                                                          \\ \hline
\multicolumn{1}{|l|}{}           & \multicolumn{1}{l|}{Prec.} & \multicolumn{1}{l|}{Recall} & \multicolumn{1}{l|}{Acc.} & \multicolumn{1}{l|}{F1}   & \multicolumn{1}{l|}{Prec.} & \multicolumn{1}{l|}{Recall} & \multicolumn{1}{l|}{Acc.} & \multicolumn{1}{l|}{F1}   & \multicolumn{1}{l|}{Prec.} & \multicolumn{1}{l|}{Recall} & \multicolumn{1}{l|}{Acc.} & \multicolumn{1}{l|}{F1}   \\ \hline
\multicolumn{1}{|l|}{credit-j}   & \multicolumn{1}{l|}{0.78}  & \multicolumn{1}{l|}{0.72}   & \multicolumn{1}{l|}{0.78} & \multicolumn{1}{l|}{0.75} & \multicolumn{1}{l|}{\textbf{0.89}}  & \multicolumn{1}{l|}{0.69}   & \multicolumn{1}{l|}{0.82} & \multicolumn{1}{l|}{0.77} & \multicolumn{1}{l|}{0.86}  & \multicolumn{1}{l|}{\textbf{0.90}}   & \multicolumn{1}{l|}{\textbf{0.89}} & \multicolumn{1}{l|}{\textbf{0.88}} \\ \hline
\multicolumn{1}{|l|}{breast-w}   & 0.92                       & 0.87                        & 0.93                      & 0.89                      & \textbf{0.98}              & 0.65                        & 0.87                      & 0.76                      & 0.94                       & \textbf{0.92}               & \textbf{0.95}             & \textbf{0.92}             \\ \hline
\multicolumn{1}{|l|}{ecoli}      & 0.85                       & 0.75                        & 0.84                      & 0.80                      & 0.95                       & 0.84                        & 0.92                      & 0.89                      & \textbf{0.95}              & \textbf{0.88}               & \textbf{0.93}             & \textbf{0.91}             \\ \hline
\multicolumn{1}{|l|}{kidney}     & 0.96                       & 0.92                        & 0.93                      & 0.94                      & \textbf{0.99}              & 0.95                        & \textbf{0.96}             & \textbf{0.97}             & 0.93                       & \textbf{0.95}               & 0.93                      & 0.94                      \\ \hline
\multicolumn{1}{|l|}{voting}     & 0.97                       & 0.94                        & 0.95                      & 0.95                      & 0.98                       & 0.95                        & 0.96                      & 0.96                      & \textbf{0.98}              & \textbf{0.96}               & \textbf{0.97}             & \textbf{0.97}             \\ \hline
\multicolumn{1}{|l|}{autism}     & 0.73                       & 0.43                        & 0.79                      & 0.53                      & \textbf{0.88}              & 0.38                        & 0.81                      & 0.52                      & 0.84                       & \textbf{0.88}               & \textbf{0.91}             & \textbf{0.86}             \\ \hline
\multicolumn{1}{|l|}{ionosphere} & 0.89                       & 0.87                        & 0.85                      & 0.88                      & 0.92                       & 0.85                        & 0.86                      & 0.88                      & \textbf{0.91}              & \textbf{0.86}               & \textbf{0.86}             & \textbf{0.89}             \\ \hline
\multicolumn{1}{|l|}{sonar}      & 0.74                       & 0.56                        & 0.66                      & 0.64                      & 0.81                       & 0.72                        & 0.74                      & 0.76                      & \textbf{0.87}              & \textbf{0.75}               & \textbf{0.78}             & \textbf{0.80}             \\ \hline
\multicolumn{1}{|l|}{heart}      & 0.76                       & 0.75                        & 0.78                      & 0.75                      & 0.79                       & 0.70                        & 0.79                      & 0.74                      & \textbf{0.82}              & 0.74                        & \textbf{0.82}             & \textbf{0.78}             \\ \hline
\multicolumn{1}{|l|}{wine}       & 0.94                       & \textbf{0.86}               & 0.93                      & 0.89                      & 0.91                       & 0.85                        & 0.92                      & 0.88                      & \textbf{0.98}              & 0.85                        & \textbf{0.93}             & \textbf{0.91}             \\ \hline \hline
\multicolumn{1}{|l|}{Average}       & 0.86                       & 0.79               & 0.85                      & 0.82                      & 0.9                       & 0.77                        & 0.87                      & 0.82                      & \textbf{0.92}              & \textbf{0.87}                        & \textbf{0.91}             & \textbf{0.89}             \\ \hline
\end{tabular}
\caption{Evaluation of Our Three Experiments with 10 UCI Datasets}
\label{tbl:accuracies}
\end{table*}

Table \ref{fig:runningTime} compares the average running time of ALEPH against LIME-FOLD. For all 10 datasets, FOLD algorithm terminates in less than one minute. All experiments were run on an Intel Core i7 CPU @ 2.7GHz with 16 GB RAM and a 64-bit Windows 10. The FOLD algorithm is a Java application that uses JPL library to connect to SWI prolog. ALEPH v.5 has been ported into SWI-Prolog by \cite{alephswiprolog}.

\begin{table}[]
\centering
\begin{tabular}{ll|l|l|}
\cline{3-4}
                                 &      & \multicolumn{2}{l|}{Running Time (s)} \\ \hline
\multicolumn{1}{|l|}{Data Set}   & size & ALEPH           & LIME-FOLD           \\ \hline
\multicolumn{1}{|l|}{credit-j}   & 125  & 1680            & 15                  \\ \hline
\multicolumn{1}{|l|}{breast-w}   & 699  & 83              & 7.8                 \\ \hline
\multicolumn{1}{|l|}{ecoli}      & 336  & 132             & 3                   \\ \hline
\multicolumn{1}{|l|}{kidney}     & 400  & 24              & 0.6                 \\ \hline
\multicolumn{1}{|l|}{voting}     & 435  & 252             & 1.8                 \\ \hline
\multicolumn{1}{|l|}{autism}     & 704  & 480             & 10.8                \\ \hline
\multicolumn{1}{|l|}{ionosphere} & 351  & 1080            & 4.8                 \\ \hline
\multicolumn{1}{|l|}{sonar}      & 208  & 834             & 9.6                 \\ \hline
\multicolumn{1}{|l|}{heart}      & 270  & 277             & 18.6                \\ \hline
\multicolumn{1}{|l|}{wine}       & 178  & 18              & 1.8                 \\ \hline
\end{tabular}
\caption{Average Running Time Comparison}
\label{fig:runningTime}
\end{table}
Table \ref{tbl:accuracies} presents the comparison of classification metrics on each of the 10 UCI datasets. The best performer is highlighted with boldface font. In 9 cases, the LIME-FOLD produces a classifier with higher $F_1$ score. However, in case of ``kidney'', LIME-ALEPH produces the highest $F_1$ score although, it generates almost twice as many clauses as LIME-FOLD does in this dataset.
\section{Related Work}
\label{sec:related}
A survey of ILP can be found in \cite{ilp20}. Rule extraction from statistical Machine Learning models has been a long-standing goal of the community. The rule extraction algorithms from machine learning models are classified into two categories: 1) Pedagogical (i.e., learning symbolic rules from black-box classifiers without opening them) 2) Decompositional (i.e., to open the classifier and look into the internals). TREPAN \cite{trepan} is a successful pedagogical algorithm that learns decision trees from neural networks. SVM+Prototypes \cite{svmplus} is a decompositional rule extraction algorithm that makes use of KMeans clustering to extract rules from SVM classifiers by focusing on support vectors. Another rule extraction technique that is gaining attention recently is ``RuleFit" \cite{rulefit}. RuleFit learns a set of weighted rules from ensemble of shallow decision trees combined with original features. In ILP community also, researchers have tried to combine statistical methods with ILP techniques. Support Vector ILP \cite{svmilp} uses ILP hypotheses as kernel in dual form of the SVM algorithm. kFOIL \cite{kfoil} learns an incremental kernel for SVM algorithm using a FOIL style specialization. nFOIL \cite{nfoil} integrates the Naive-Bayes algorithm with FOIL. The advantage of our research over all of the above mentioned research work is that, first it is model agnostic, second it is scalable thanks to the greedy nature of our clause search. 

\section{Conclusions and Future Work}
In this paper we presented a heuristic based algorithm called LIME-FOLD. This novel algorithm can induce very concise nonmonotonic logic programs to explain the implicit rules captured by any sophisticated classifier such as XGBoost. LIME is used to provide explanations as to what features contribute most to the XGBoost model's prediction. However, these explanations are local and specific to a given data sample and do not represent the global model's behavior. We have shown that by filtering out the locally irrelevant features, a transformed dataset is created that, once given to the LIME-FOLD algorithm, yields a very concise nonmonotonic logic program that is much more accurate than hypotheses induced by ALEPH, a state-of-the-art ILP system. We have justified our claim by running LIME-FOLD and ALEPH on the standard and transformed UCI standard benchmarks. In terms of the running time, our LIME-FOLD algorithm in average runs 80 times faster than ALEPH on 10 datasets reported in this paper.

There are number of directions for future work: (i) Explanations provided by LIME could be used to discover the sub-concepts; for example, one may create vectors from the explanations to examine whether a clustering algorithm such as KMeans could successfully separate the sub concepts. (ii) While LIME perturbs each feature in isolation from other features, the interactions between features are ignored. ILP on the other hand is a promising technique when it comes to feature construction. We plan to investigate how ILP and LIME can mutually reinforce each other to produce better machine learning algorithms.

\section*{Acknowledgements}
Authors are partially supported by NSF Grant IIS 1718945. We would like to cordially thank Dr. Gautam Das for bringing LIME to our attention.

\bibliographystyle{splncs03}
\bibliography{mycitations}
\end{document}